\documentclass[sigconf, nonacm]{acmart}
\usepackage{pifont}
\usepackage{wrapfig}
\usepackage{tabularx}
\usepackage{adjustbox}
\usepackage{svg}

\usepackage{float}
\usepackage{array}
\usepackage{multirow}
\usepackage{enumitem}
\newcommand\vldbyear{2024}
\newcommand\vldbworkshop{LLM+KG}
\newcommand\vldbauthors{\authors}
\newcommand\vldbtitle{\shorttitle}
\newcommand\vldbavailabilityurl{
}
\newcommand\vldbpagestyle{plain}

\begin{document}

\title{Research Trends for the Interplay between Large Language Models and Knowledge Graphs}

\author{Hanieh Khorashadizadeh}
\affiliation{%
  \institution{University of L\"ubeck}
  \city{L\"ubeck}
  \state{Germany}
}
\email{hanieh.khorashadizadeh@uni-luebeck.de}

\author{Fatima Zahra Amara}
\affiliation{
  \institution{University of Khenchela}
  \streetaddress{}
  \city{Khenchela}
  \country{Algeria}
}
\email{f.amara@univ-khenchela.dz}

\author{Morteza Ezzabady}
\affiliation{%
  \institution{Universit\'e de Toulouse}
  \city{Toulouse}
  \country{France}
}
\email{morteza.ezzabady@irit.fr}

\author{Fr\'ed\'eric Ieng}
\affiliation{%
  \institution{Universit\'e Paris Cit\'e}
  \city{Paris}
  \country{France}
}
\email{frederic.ieng@u-paris.fr}

\author{Sanju Tiwari}
\affiliation{%
  \institution{Alliance University}
  \city{Bangalore}
  \country{India}
}
\email{tiwarisanju18@ieee.org}

\author{Nandana Mihindukulasooriya}
\affiliation{%
  \institution{IBM Research}
  \city{New York}
  \country{US}
}
\email{nandana@ibm.com}

\author{Jinghua Groppe}
\affiliation{%
  \institution{University of L\"ubeck}
  \city{L\"ubeck}
  \country{Germany}
}
\email{jinghua.groppe@uni-luebeck.de}

\author{Soror Sahri}
\affiliation{%
  \institution{Universit\'e Paris Cit\'e}
  \streetaddress{}
  \city{Paris}
  \country{France}
}
\email{soror.sahri@parisdescartes.fr}

\author{Farah Benamara}
\affiliation{%
  \institution{Universit\'e de Toulouse}
  \streetaddress{}
  \city{Toulouse}
  \country{France}
}
\email{farah.benamara@irit.fr}

\author{Sven Groppe}
\orcid{0000-0001-5196-1117}
\affiliation{%
  \institution{University of L\"ubeck}
  \city{L\"ubeck}
  \country{Germany}
}
\email{sven.groppe@uni-luebeck.de}
\begin{abstract}
This survey investigates the synergistic relationship between Large Language Models (LLMs) and Knowledge Graphs (KGs), which is crucial for advancing AI's understanding, reasoning, and language processing capabilities. It aims to address gaps in current research by exploring areas such as KG Question Answering, ontology generation, KG validation, and enhancing KG accuracy and consistency through LLMs. The paper further examines the roles of LLMs in generating descriptive texts and natural language queries for KGs. Through a structured analysis that includes categorizing LLM-KG interactions, examining methodologies, and investigating collaborative uses and potential biases, this study seeks to provide new insights into the combined potential of LLMs and KGs. It highlights the importance of their interaction for improving AI applications and outlines future research directions.
\end{abstract}

\maketitle

\pagestyle{\vldbpagestyle}
\begingroup\small\noindent\raggedright\textbf{VLDB Workshop Reference Format:}\\
\vldbauthors. \vldbtitle. VLDB \vldbyear\ Workshop: \vldbworkshop.\\ 
\endgroup
\begingroup
\renewcommand\thefootnote{}\footnote{\noindent
This work is licensed under the Creative Commons BY-NC-ND 4.0 International License. Visit \url{https://creativecommons.org/licenses/by-nc-nd/4.0/} to view a copy of this license. For any use beyond this license, obtain permission by emailing \href{mailto:info@vldb.org}{info@vldb.org}. Copyright is held by the owner/author(s). Publication rights licensed to the VLDB Endowment. \\
\raggedright Proceedings of the VLDB Endowment. 
ISSN 2150-8097. \\
}\addtocounter{footnote}{-1}\endgroup

\ifdefempty{\vldbavailabilityurl}{}{
\vspace{.3cm}
\begingroup\small\noindent\raggedright\textbf{VLDB Workshop Artifact Availability:}\\
The source code, data, and other artifacts have been made available at \url{\vldbavailabilityurl}.
\endgroup
}

\section{Introduction}

In the era of artificial intelligence, Large Language Models (LLMs) and Knowledge Graphs (KGs) stand out as two pivotal technologies driving advancements in machine understanding, reasoning, and natural language processing. LLMs, such as OpenAI's GPT series~\cite{bang2023multitask}, have demonstrated remarkable capabilities in generating human-like text, answering questions, and even creating content across various domains~\cite{khorashadizadeh2023exploring}. On the other hand, KGs organize and integrate information in a structured format, enabling machines to understand and infer relationships between entities in the real world \cite{xu2021enabling}. The synergy between LLMs and KGs opens new frontiers for AI applications, enhancing machines' ability to process, interpret, and generate information with context and accuracy. Our intention with this survey is to fill the gaps left by previous survey papers \cite{pan2024unifying,pan2023large,hu2023survey,yang2024facts}, offering fresh perspectives and uncovering new frontiers in the interaction between LLMs and KGs. For instance, our analysis delves deeper into KG Question Answering, encompassing multi-hop question answering, multi-hop question generation, and KG-powered chatbots.
Additionally, our study extends to KG validation, an area not extensively discussed in previous survey papers. Table~\ref{tab:PreviousSurveys} outlines the categories discussed in previous research papers and this survey paper. Given the transformative potential and challenges of integrating LLMs with KGs, this paper seeks to explore the following research questions:

\renewcommand{\labelenumi}{\theenumi.}
\begin{enumerate}[topsep=0em, partopsep=0em
]
\item How can LLMs generate descriptive textual information for entities in a KG?
\item How can we employ LLMs in ontology generation?
\item How can LLMs help in detecting inconsistencies within KGs?
\item How can LLMs improve the accuracy, consistency, and completeness of KGs through fact-checking?
\item How can LLMs contribute to providing accurate answers for KG Question Answering?
\item How can LLMs effectively generate queries from natural language texts? (Text to Sparql or Cypher)
\end{enumerate}

\begin{figure*}
   \centering
\includegraphics[width=\textwidth]{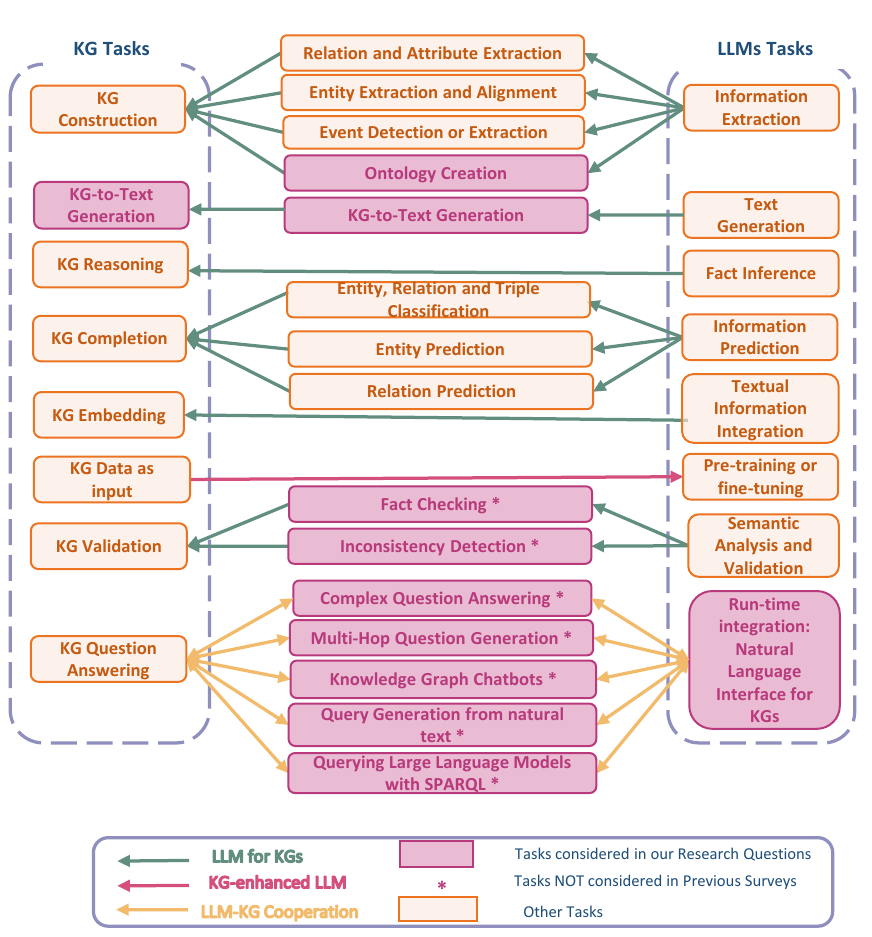}
    \caption{Categorization of the interplay between LLMs and KGs}
    \label{fig:categorization}
\end{figure*}

The interplay between LLMs and KGs can be categorized into three distinct types. The first, LLMs for KGs, demonstrates how LLMs are utilized to bolster and refine the functionality of KGs. The second, KG-enhanced LLMs, shows the inverse, where KGs are leveraged to enhance and inform the capabilities of LLMs. Lastly, LLM-KG Cooperation is depicted, representing a collaborative and synergistic relationship where LLMs and KGs work together to achieve more advanced and complex outcomes. Figure~\ref{fig:categorization} illustrates the categorization of the interaction between LLMs and KGs. The remainder of this paper is structured to include an in-depth exploration of the three distinct types along with their respective sub-categories. The primary focus of this research is to address our formulated research questions. These specific areas of inquiry are highlighted in Figure~\ref{fig:categorization}, distinctly marked in pink, mentioned as Tasks Considered in our Research Questions. Those sections highlighted with stars represent topics that were not addressed in previous survey papers.
The paper is organized as follows: Section~\ref{sec:LLMforKG} explores the role of LLMs within the context of KGs and Section~\ref{sec:KG-enhancedLLM} investigates how KGs can enhance LLMs. In Section~\ref{sec:LLM-KGCooperation}, we examine the collaborative utilization of LLMs and KGs. Section~\ref{sec:OpenChallenges} entails a statistics analysis and the identification of open challenges. Finally, we conclude in Section~\ref{sec:Conclusions}.

\begin{table*}[ht]
\caption{Categorizations addressed by previous survey papers}
\label{tab:PreviousSurveys}
\centering
\normalsize 
\adjustbox{max width=\textwidth}{
\begin{tabularx}{\textwidth}{|>{\centering\arraybackslash}p{4cm}|>{\centering\arraybackslash}p{6.2cm}|>{\centering\arraybackslash}p{1cm}|>{\centering\arraybackslash}p{1cm}|>{\centering\arraybackslash}p{1cm}|>{\centering\arraybackslash}p{1cm}|>{\centering\arraybackslash}p{1cm}|}
\hline
\textbf{Main Category} & \textbf{Subcategory} & \textbf{Survey: \cite{pan2024unifying}} & \textbf{Survey: \cite{pan2023large}} & \textbf{Survey: \cite{hu2023survey}} & \textbf{Survey: \cite{yang2024facts}} & \textbf{\textit{Our Survey}} \\
\hline
\multirow{4}{=}{\centering \textbf{KG Construction}} & \textbf{Relation and Attribute Extraction} & \ding{51} & \ding{51} & \ding{55} & \ding{55} & \ding{51} \\
\cline{2-7}
 & \textbf{Entity Extraction and Alignment} & \ding{51} & \ding{51} & \ding{55} & \ding{55} & \ding{51} \\
\cline{2-7}
 & \textbf{Event Detection or Extraction} & \ding{55} & \ding{55} & \ding{55} & \ding{55} & \ding{55} \\
\cline{2-7}
 & \textbf{Ontology Creation} & \ding{55} & \ding{51} & \ding{55} & \ding{55} & \ding{51} \\
\hline
\multirow{1}{=}{\centering \textbf{KG-to-Text Generation}} & \textbf{KG-to-Text Generation} & \ding{51} & \ding{55} & \ding{55} & \ding{55} & \ding{51} \\
\hline
\multirow{1}{=}{\centering \textbf{KG Reasoning}} & \textbf{KG Reasoning} & \ding{51} & \ding{51} & \ding{55} & \ding{55} & \ding{51} \\
\hline
\multirow{3}{=}{\centering \textbf{KG Completion}} & \textbf{Entity, Relation and Triple Classification} & \ding{51} & \ding{51} & \ding{55} & \ding{55} & \ding{51} \\
\cline{2-7}
 & \textbf{Entity Prediction} & \ding{51} & \ding{51} & \ding{55} & \ding{55} & \ding{51} \\
\cline{2-7}
 & \textbf{Relation Prediction} & \ding{55} & \ding{51} & \ding{55} & \ding{55} & \ding{51} \\
\hline
\multirow{1}{=}{\centering \textbf{KG Embedding}} & \textbf{KG Embedding} & \ding{51} & \ding{55} & \ding{55} & \ding{55} & \ding{51} \\
\hline
\multirow{1}{=}{\centering \textbf{KG-enhanced LLM}} & \textbf{KG-enhanced LLM} & \ding{51} & \ding{51} & \ding{51} & \ding{51} & \ding{51} \\
\hline
\multirow{2}{=}{\centering \textbf{KG Validation}} & \textbf{Fact Checking} & \ding{55} & \ding{55} & \ding{55} & \ding{55} & \ding{51} \\
\cline{2-7}
 & \textbf{Inconsistency Detection} & \ding{55} & \ding{55} & \ding{55} & \ding{55} & \ding{51} \\
\hline
\multirow{5}{=}{\centering \textbf{KG Question Answering}} & \textbf{Complex Question Answering} & \ding{55} & \ding{55} & \ding{55} & \ding{55} & \ding{51} \\
\cline{2-7}
 & \textbf{Multi-Hop Question Generation} & \ding{55} & \ding{55} & \ding{55} & \ding{55} & \ding{51} \\
\cline{2-7}
 & \textbf{Knowledge Graph Chatbots} & \ding{55} & \ding{55} & \ding{55} & \ding{55} & \ding{51} \\
\cline{2-7}
 & \textbf{Query Generation from natural text} & \ding{55} & \ding{55} & \ding{55} & \ding{55} & \ding{51} \\
\cline{2-7}
 & \textbf{Querying Large Language Models with SPARQL} & \ding{55} & \ding{55} & \ding{55} & \ding{55} & \ding{51} \\
\hline
\end{tabularx}
}
\end{table*}

\section{LLM for KG}\label{sec:LLMforKG}
\subsection{KG Construction}

\subsubsection{Ontology Creation}
This section of the paper explicitly addresses Research Question 2, delving into the application of LLMs in ontology generation for KGs. LLMs exhibit significant promise in constructing ontologies \cite{neuhaus2023ontologies}, where the organized representation of knowledge is essential for facilitating intelligent information retrieval and reasoning.
LLMs contribute to creating and enriching comprehensive ontologies, with domain-specific fine-tuning enhancing their integration into specialized areas. While constructing a KG, \cite{Ezzabady2024KGLLM} built an ontology for COVID-19 and demonstrated the effectiveness of LLMs.
LLMs are promising for ontology generation but face computational intensity and bias challenges. They require broader evaluation yet remain valuable in ontology creation and intelligent systems.  

\paragraph{Concept and Relation Extraction:}
LLMs like GPT-3, with their natural language understanding abilities, are suitable for extracting ontology concepts and relations. Concept extraction is the process of identifying and extracting concepts from text. The process involves extracting linguistic realizations of domain-specific concepts and accumulating semantic term variations with sense disambiguation and domain-specific synonym identification \cite{schaeffer2023olaf}. Relation extraction aims to identify non-taxonomic relations between concepts \cite{babaei2023llms4ol}. It seeks to identify relations between text entities and can be enhanced by using techniques like Named Entity Recognition (NER) and additional context provided by LLMs \cite{caufield2023structured,chang2023concept}.

\paragraph{Property Identification:}
Contributions like \cite{strakova2023extending} investigate using fine-tuned LLMs for pre-annotating data in ontology property identification, helping annotators efficiently discern relevant properties and their correlation with automated scores, thereby reducing annotation times.

\paragraph{Ontology Enrichment:}
LLMs can help to improve current ontologies by adding new information or improving concept definitions. This is especially beneficial when working with dynamic domains in which the ontology must adapt to new facts. Advanced machine learning methods, remarkably fine-tuned LLMs, have been researched for data pre-annotation and ontology enrichment~\cite{strakova2023extending}.

\paragraph{Ontology Alignment:}
LLMs can help to align ontologies by understanding and comparing the semantics of ideas and relationships among ontologies. Ontology alignment with LLMs has been studied in the literature. One method suggests a neurosymbolic design that combines the flexibility of LLMs and the domain orientation of Enterprise KGs (EKGs) \cite{baldazzi2023fine}. Another work investigates the use of fine-tuned LLMs to pre-annotate data in a lexical extension job, specifically adding descriptive words to an existing ontology~\cite{strakova2023extending}. 

\paragraph{Text-to-Ontology Mapping:}
Train LLMs to map natural language text to specific ontology concepts and relationships. This can be accomplished using supervised learning, which involves training the model using annotated instances of text-to-ontology mappings. Text-to-ontology mapping using LLMs has been explored in multiple papers. One approach involves using artificial neural networks and classifiers to match texts to relevant ontologies \cite{korel2023text}.

\paragraph{Ontology Learning:}
Ontology learning automates the construction of ontologies by identifying concepts from text, aiming for a cost-effective, scalable knowledge representation across disciplines. Recent advances in LLMs, such as BERT, offer a promising alternative to traditional, time-consuming ontology creation, outperforming classic word embedding models in various nlp tasks and providing a more efficient method for knowledge acquisition in ontology development \cite{schaeffer2023olaf,chen2023contextual,babaei2023llms4ol,funk2023towards}.


\subsubsection{Entity Extraction and Alignment:}
The Named Entity Recognition (NER) task involves being provided with a sentence, and the objective is to predict the set of entities present within that sentence\cite{ashok2023promptner}. 
Some papers concentrated on prompting to extract entities. \citeauthor{ashok2023promptner} approach named PromptNER, comprises four essential components: a backbone LLM, a prompt outlining the set of entity types, a small set of examples from the target domain, and a specific format for outputting the extracted entities. Some approaches employ instruction-tuning. It is a technique in which pre-trained autoregressive language models undergo finetuning to adhere to natural language instructions and produce responses \cite{ouyang2022training}. \citeauthor{zhou2023universalner} proposed a focused distillation method with instruction tuning tailored to a specific mission.

\subsubsection{Relation Extraction:}
LLMs have notably advanced the complex task of relation extraction between text entities, revolutionizing the identification and interpretation of semantic connections in textual data. In particular, GPT-RE \cite{wan_gpt-re_2023} introduced a novel approach by infusing task-specific knowledge and logic-enriched demonstrations. This approach effectively reduces the low relevance issues between entity-relation pairings and overcomes the limitations in explaining input-label correlations. Furthermore, considering the vast array of predefined relation types and the challenges posed by uncontrolled LLMs, the study \cite{li_semi-automatic_2023} suggests incorporating a natural language inference module within LLMs. This integration focuses on generating relation triples to greatly improve the usefulness and precision of document-level relation datasets. These advancements in relation extraction demonstrate LLMs' capability to improve semantic analysis in text, and we have organized these methods by their distinct learning approaches. This classification comprises:

\paragraph{Supervised Fine-tuning:} Supervised fine-tuning known for being prevalent and promising, involves extensively inputting training data to fine-tune LLMs. It allows LLMs to learn and comprehend the structural patterns in data, enhancing their ability to generalize.
The Deepstruct model, referenced by \cite{wang_deepstruct_2023}, uses structure pre-training on diverse corpora to improve language models' understanding of complex language structures, leading to a more nuanced interpretation of linguistic patterns.
\citeauthor{huguet_cabot_rebel_2021} enhanced relation extraction by training a specialized BART-style model \cite{lewis_bart_2020} with a novel triplet linearization technique, which demonstrates the importance of supervised fine-tuning in advancing LLMs for complex tasks like relation extraction.

\paragraph{Few-shot learning:} Few-shot learning, which relies on a limited number of labeled examples, faces significant challenges in machine learning. The lack of data can cause problems like overfitting and difficulty in understanding complex data relationships, as pointed out by \cite{huang2020few}. 
The advent of LLMs has been a significant breakthrough in overcoming few-shot learning challenges. 
\citeauthor{xu_how_2023} introduced two strategies for using LLMs in relation extraction: 1) in-context learning (ICL) and 2) data generation. ICL involves using detailed prompts with task definitions and labels to help LLMs understand relation extraction nuances, which is particularly useful in few-shot settings with limited data. To further address data scarcity, they used LLMs to generate additional training material by creating prompts with instance descriptions and examples. In a complementary study, \cite{wadhwa_revisiting_2023} examined the effectiveness of LLMs in few-shot relation extraction through in-context learning. They discovered that LLMs could equal state-of-the-art techniques using just a few examples. Their method included enhancing relation extraction labels with Chain of Thought (CoT) explanations from GPT-3 and fine-tuning the Flan-T5 model with this enriched data. 

\paragraph{Zero-shot learning:} Zero-shot learning aims to empower models to generalize to new tasks and domains beyond their training by adapting the pre-training of LLMs to unfamiliar situations. 
The extensive knowledge embedded in LLMs demonstrates their remarkable capability for zero-shot tasks in various fields. Key studies such as those by \cite{kojima_large_2023,wei_zero-shot_2023}, and recent advancements 
\cite{li_revisiting_2023}, exemplify the potential of LLMs in adapting to tasks they were not explicitly trained for. Adding to this discourse, \cite{yuan_zero-shot_2023} critically assessed ChatGPT's zero-shot learning capability in temporal relation extraction. The study emphasized ChatGPT's skill in grasping complex temporal relations but also noted its limitations in consistency and handling long-dependency relations. 

\subsection{KG-to-Text Generation:}
This segment of the paper is dedicated to addressing Research Question 1, which investigates the effective use of LLMs in generating descriptive textual information for entities within KGs. KG-to-Text generation encompasses the procedure of transforming structured data held within a KG into human-readable, natural-language text \cite{colas2022gap}.
\citeauthor{colas2022gap} combined a pre-trained LM with graph attention information by modifying the encoder layer. Initially, they convert the KG into a textual string by linearizing its structure. Then, the vector representation becomes contextualized through the incorporation of graph attention data.
Several methods involve the fine-tuning of pre-trained language models(PLM), specifically on KG-to-text datasets, to address this particular task \cite{ribeiro2020investigating}. \citeauthor{chen2020kgpt} gathered a corpus named KGTEXT, which involves matching linked sentences sourced from Wikipedia with knowledge subgraphs obtained from WikiData and then conducting pre-training and subsequently fine-tuning on KG-to-Text datasets.
\citeauthor{ke2021jointgt} developed a basic yet effective structure-aware semantic aggregation module that integrates into every Transformer layer within the encoder. The purpose of this module is to maintain the structural details present in the input graphs. Then followed by three pre-training tasks aimed at explicitly acquiring graph-text alignments.
\citeauthor{li2021few} investigated few-shot KG-to-text generation, which generates the output with the help of around several hundred instances. Initially,  their model aligns KG representations, encoded by graph neural networks, PLM-based entity representations, aiming to close the semantic gap. It then proposes a specialized strategy, called relation-biased breadth-first search (RBFS), to arrange the KG into a well-structured entity sequence for integration into PLMs. Lastly, the model employs multi-task learning, simultaneously training the main text generation and an auxiliary KG reconstruction task. This approach significantly improves the semantic alignment between the input KG and the resulting text, allowing the model to generate accurate descriptions of the KG.

\subsection{KG Reasoning}
KG Reasoning involves deriving new facts from existing ones using logical inference. Recently, there has been a growing emphasis on exploring KG reasoning due to its potential to derive new knowledge and conclusions from existing data\cite{chen2020review}. It has applications for KG completion and KG question answering~\cite{chen2020review}. \citeauthor{choudhary2023complex} developed Language-guided Abstract Reasoning over Knowledge graphs (LARK) model that is designed to harness the reasoning power of LLMs for efficiently answering first-order logic (FOL) queries on knowledge graphs. LARK identifies relevant subgraph contexts within abstract KGs by utilizing the entities and relationships in the queries. It then conducts chain reasoning over these contexts using prompts from LLMs to break down and process logical queries. \citeauthor{luo2023reasoning} introduced Reasoning on Graphs (RoG), which combines LLMs and KGs to perform reliable and easily understandable reasoning. They proposed a framework for planning-retrieval-reasoning to tackle challenges such as hallucinations and knowledge deficiencies. Initially, the planning module produces relation paths in the form of faithful plans using KGs, forming the basis for the RoG approach. Subsequently, these devised plans are employed to extract legitimate paths of reasoning from KGs to carry out reliable reasoning through the retrieval-reasoning module. \citeauthor{kim2023kg} introduced KG-GPT, a versatile framework for reasoning on KG that operates in three stages: Sentence Segmentation, Graph Retrieval, and Inference. These stages are designed to break down sentences, extract relevant components from the graph, and draw logical conclusions.

\subsection{KG Completion}
In KG completion, the main tasks are (i) triple classification, validating the accuracy of a given triple; (ii) link prediction for predicting missing components like the head, tail, or relation in a triple. And (iii) entity classification for categorizing entities, also known as determining the entity's type~\cite{alam2023towards}. The triple-based methods relying on structural information are widely used for KG completion, mostly the embedding methods~\cite{bordes2013translating,lin2015learning,trouillon2016complex}. Several works have attempted to prove more effectiveness than triple-based completion by extracting factual knowledge from LMs embedding spaces instead of simple structures. Examples of these text-based methods are KG-BERT \cite{yao2019kg} and the fine-tuning of GPT-2~\cite{Biswas2021ContextualLM}. KG-BERT~\cite{yao2019kg} is a leading example of applying LLMs to KG completion. It fine-tunes for this task by treating triples as textual sequences.
The model uses KG-BERT to score sequences of entity-relation-entity, representing these elements through their names or descriptions. These sequences of words are then used as input for fine-tuning the BERT model. KG-BERT was identified with two shortcomings: 1) the model's struggle to learn relational data in KGs, and 2) difficulty in identifying the correct answers among similar candidates. To improve this, multi-task learning methods combining relation prediction and relevance ranking with link prediction were proposed, as in \cite{kim2020multi}. This enhanced model better grasps KG relationships and effectively handles lexical similarities. In another study, \cite{wang2021structure} introduced the Structure-augmented Text Representation (StAR) model aimed at improving link prediction in KG completion. The model begins with a Siamese-style textual encoder applied to triples, generating two contextualized representations. A distinctive scoring module using dual strategies captures both contextualized and structured knowledge, enhanced by a self-adaptive ensemble scheme with graph embedding techniques to boost performance. \citeauthor{wang2022simkgc} presented SimKGC, a straightforward approach to enhance text-based KG completion. It recognizes the crucial aspect of conducting efficient contrastive learning. Building on recent advancements in contrastive learning. SimKGC utilizes a bi-encoder architecture and incorporates three types of negative examples to achieve its objectives. 
Sequence-to-sequence PLM methods have also shown state-of-the-art performance by redefining the KG completion challenge as a sequence-to-sequence generation task as demonstrated in KG-S2S~\cite{chen2022knowledge} and GenKGC~\cite{xie2022discrimination}. GenKGC incorporates innovative techniques, such as relation-guided demonstration and entity-aware hierarchical decoding, aimed at improving the quality of representation learning and accelerating the inference process. 
Most of these recent text-based methods require specific fine-tuning for KGs, which limits their efficiency. New training-free methods with no fine-tuning were recently proposed to address this limitation, as in KICGPT \cite{wei2023kicgpt}.

\subsection{KG Embedding}
KG Embedding involves mapping elements of a KG into vectors in a continuous vector space.
Unlike Graph Neural Networks, which learn KG embeddings from data explicitly represented by nodes and edges, LM learns KG embeddings from natural language. Given a piece of text, LMs learn representations of entities, which capture context information containing relations of entities. The entity relations are found by an attention mechanism, which can create an association between two entities independent of their location. The more text an LM sees the more entities and their relationships the LM can learn; the larger an LM, the more contextual information the representation contains. LLMs contain billions of parameters and are trained over a huge amount of textual data (e.g. 45 TB for GPT-3). LLMs perfectly meet these requirements, and so are a powerful tool for learning KG embeddings. LLMs' ability to KG embedding attracts researchers to use them directly to conduct KG tasks, like entity alignment~\cite{lippolis2023enhancing}. We can also use the representation of entities learned by LLMs in the small-sized models, and this should significantly reduce the amount of training data needed and the time of training and, meanwhile, create well-generalized models. An extensive experiment is needed to investigate the efficiency of applying embeddings of LLMs into small-sized models for KG analysis tasks. 

\subsection{KG Validation}
\subsubsection{Fact Checking}
In this section, we address Research Question 4, which is centered on how LLMs can enhance the accuracy, consistency, and completeness of KGs through the process of fact-checking. The advent of LLMs has the potential to significantly impact the fight against misinformation, presenting both opportunities and challenges. LLMs, with their extensive world knowledge and strong reasoning abilities, offer promise in combating misinformation. However, the flip side is that these models can also be easily manipulated to generate deceptive misinformation at scale \cite{chen2023combating}. 
An approach to address misinformation in a KG involves verbalizing each triple within the KG and prompting LLMs to identify instances of misinformation \cite{chen2023can,bang2023multitask}. To address the limitations of potentially outdated or insufficient knowledge in LLMs for detecting factual errors, some research has explored augmenting LLMs with external knowledge~\cite{cheung2023factllama} or utilizing tools~\cite{chern2023factool} for misinformation detection. The methodologies developed for factual checking in LLMs can be applied to KG fact-checking by converting KG triples into textual representations.
\subsubsection{Inconsistency Detection}
This section delves into Research Question 3, examining the role of LLMs in identifying and fixing inconsistencies in KGs.
KG consistency is about ensuring that the information in a KG is not self-contradictory~\cite{zaveri2016quality}. KG consistency differs from KG accuracy in that it prioritizes the logical soundness of the KG rather than the factual correctness of individual triples. A KG might contain outdated yet logically coherent information, maintaining high consistency even with low accuracy. Conversely, high accuracy usually implies good consistency, as factual data tends to be logically harmonious~\cite{wang2021knowledge}.
Current rule-based techniques for KG completion and inconsistency detection primarily utilize structural information, overlooking the semantics of relations.
Using LLMs can harness both the semantic and structural information from KGs to generate meaningful rules. Among recent works leveraging LLMs for rule-based methods, there is ChatRule \cite{Luo2023ChatRuleML}, a framework that uses both the semantic and structural information of KGs to prompt LLMs to generate logical rules.

\section{KG-enhanced LLM}\label{sec:KG-enhancedLLM}

K-BERT~\cite{liu2020k} proposed to inject relevant domain knowledge from a KG into a sentence using a knowledge layer before the embeddings are calculated. The authors show that it improves performance in many NLP tasks. Sem-K-BERT~\cite{xu2021enabling} further extended this by integrating semantic role labeling (SRL) and by introducing semantic correlation calculation to reduce the noise. KCF-NET~\cite{DBLP:journals/access/GongLYH20} demonstrated that incorporating KG representations with context helped to improve the LM performance in machine reading comprehension. \citeauthor{DBLP:journals/access/JiDPS20} proposes a concept-enhanced pertaining approach by adding concept semantics to sentences from external sources such as KGs. Dict-BERT~\cite{DBLP:conf/acl/00020FYWX0022} leverages definitions of rare words from dictionaries to improve the performance of various NLP tasks.
Retrieval Augmented Generation (RAG) was developed to address issues such as hallucination, insufficient domain-specific knowledge, and outdated information by incorporating relevant external knowledge into the LLM prompt \cite{gao2023retrieval}. RAG has three steps: Retrieval, Generation, and Augmentation. Over time, RAG has developed into three distinct models: Naive RAG, Advanced RAG, and Modular RAG. Naive RAG operates through a three-step process: indexing, retrieval, and generation. During the indexing step, text data is divided into small segments, each of which is encoded into vector form using embedding models. In the retrieval phase, the user's query is similarly converted into a vector. The system then computes the similarity between the query vector and the vectors of the text segments, selecting the top k segments as the context for the response. Finally, in the generation step, both the query and the chosen contexts are input into an LLM as a prompt to generate a response. The Modular RAG has the ability to retrieve pertinent information from knowledge graphs\cite{gao2023retrieval}. \cite{wang2023knowledgpt} introduced KnowledgeGPT, a system that enhances LLMs with data from knowledge bases. To respond to a user query, their method generates a search code and executes it to retrieve information from the knowledge base. The output is then given to the LLM to respond to the user query. While RAG is effective in responding to queries directly related to specific excerpts from the source data where the retrieval is explicit, it may struggle with more general queries. For example, it might not perform well when asked, "What are the main points of the dataset?" To overcome this limitation, Graph RAG was introduced \cite{edge2024local}. This approach generates a summary over a knowledge graph that has been constructed by the LLM from the source data.

\begin{figure*}
    \centering
\includegraphics[width=\textwidth]{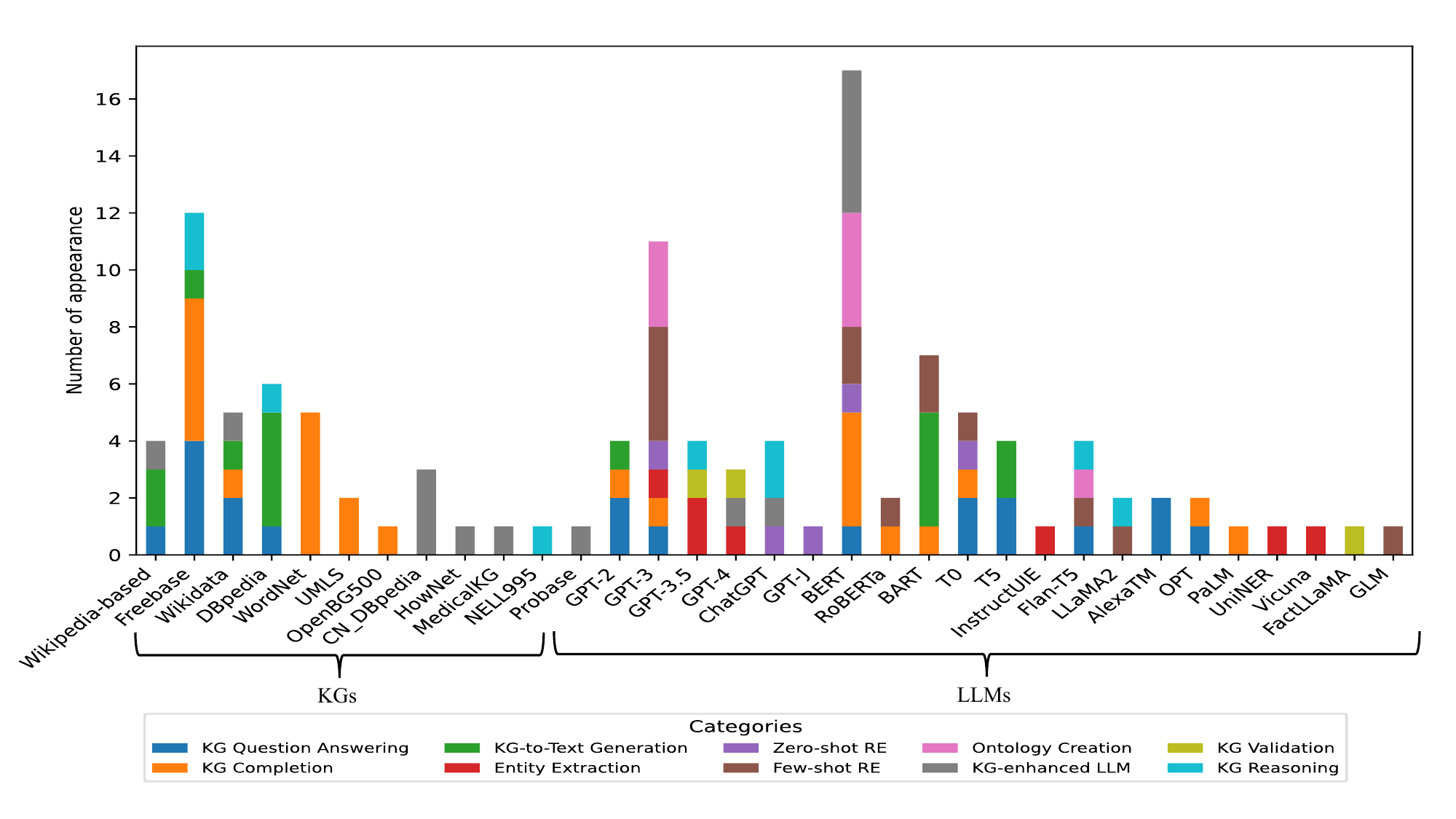}
    \caption{Statistics of the usage of LLMs and KGs in cited papers per category}
    \label{fig:LLM-KG used}
\end{figure*}

\section{LLM-KG Cooperation}\label{sec:LLM-KGCooperation}
\subsection{KG Question Answering}
This section covers research questions 5 and 6. 
\subsubsection{Multi-Hop Question Generation:} The goal of multi-hop question generation is to formulate questions demanding advanced reasoning over multiple sentences, including their associated answers \cite{li2023multi}. \citeauthor{li2023multi} created a KG Enhanced Language Model (KGEL), aiming to replicate human-like reasoning for multi-hop questions. Their method for generating multi-hop questions reflects the way humans approach complex questioning. Their strategy encompasses three key phases: (1) Gaining an understanding of the context by employing a pre-trained GPT-2 language model, (2) Merging information and reasoning through a sophisticated KG and an answer-aware dynamic Graph Attention Network (GAT), and (3) Executing the question generation using a multi-head attention generation module that capitalizes on improved latent representations. \citeauthor{aigo2021question} research primary emphasis revolved around formulating inquiries by employing KGs in conjunction with the T5 language model. They harnessed the language model's capabilities to generate questions while also applying a technique to obstruct self-attention in the encoder, all to train the model with a deliberate emphasis on maintaining the inherent structure of the graph. They harnessed the language model’s capabilities to generate questions while also applying a technique to obstruct self-attention in the encoder, all to train the model with a deliberate emphasis on maintaining the inherent structure of the graph. Their research didn’t target multi-hop question generation.

\subsubsection{Complex or Multi-hop Question Answering:}
Complex question answering over KGs involves multiple subjects, conveys interconnected relations, includes numerical operations \cite{lan2021survey}, and demands multi-hop reasoning~\cite{yang2018hotpotqa,etezadi2023state}. \citeauthor{cao2023relmkg} introduced a novel method, ReLMKG, designed for the multi-hop Knowledge Base Question Answering (KBQA) task. The approach leverages both the implicit knowledge embedded in a pre-trained language model and the explicit knowledge from KGs. They transferred the KG into a textual format, aligning it with the question in a shared space. The textualized graph and the question are input for the pre-trained language model, which acts as the question-path encoder. The lack of the inherent topological arrangement in the textual KG, which holds essential explicit knowledge for model reasoning, led to the creation of a path-centric Graph Neural Network (GNN) as the reasoning element. The instructions for the reasoning module are derived from the outputs of the pre-trained language model.
\citeauthor{sen2023knowledge} introduced an innovative approach to address queries, employing a KGQA model to extract information from a KG. Simultaneously, a language model is employed to analyze the question and facts, enabling the derivation of a well-reasoned answer. \citeauthor{baek2023knowledge}  utilized a similarity metric between KG facts and questions. This approach was employed to retrieve pertinent facts, subsequently enhancing the prompt of a language model.

\subsubsection{Query Generation from text:}
This portion of the paper specifically addresses Research Question 6, which examines the effectiveness of LLMs in generating natural language queries. It particularly focuses on converting textual descriptions into query languages like SPARQL or Cypher for KGs.
Constructing SPARQL queries from natural language questions is challenging, as it demands an understanding of both the question itself and the underlying patterns within the KG \cite{rony2022sgpt}. \citeauthor{rony2022sgpt} introduced a novel embedding technique called SGPT, tailored to encode linguistic features from the question, and suggested training methods that make use of a pre-trained language model to produce a SPARQL query. \citeauthor{kovriguina2023sparqlgen} proposed a one-shot generative approach, termed SPARQLGEN, which aims to generate SPARQL queries by augmenting LLMs with pertinent context provided in a single prompt. This prompt encompasses the question, an RDF subgraph necessary for answering the question, and an example of a correct SPARQL query for a different question.
\citeauthor{pliukhin2023improving} devised a method that incorporates diverse data sources in the SPARQL generation prompt, encompassing the question, an RDF subgraph essential for answering the question, and an illustration of a correct SPARQL query. Their method suggests a couple of enhancements to the SPARQLGEN approach.

\subsubsection{Querying LLMs with SPARQL:}
\citeauthor{saeed2023querying} explored the possibility of querying LLMs with SQl, initiating new directions for research exploration. While SQL is a powerful tool for structured datasets with defined schemas, it falls short in immediately handling information presented as unstructured text. The statement discusses an illustration of a Database-First (DB-first) approach that harnesses the capabilities of LLMs along with traditional Database Management Systems (DBMSs). This combination is utilized to establish a hybrid query execution environment, suggesting an integration of LLMs with conventional database systems for enhanced performance. This concept might also align with querying LLMs with SPARQL, which might obtain hidden relations in unstructured data.

\subsubsection{KG Chatbots:}
\citeauthor{omar2023chatgpt} conducted a detailed examination and comparison between conversational AI models like ChatGPT and traditional Question-Answering Systems (QASs) designed for KGs and suggested merging the attributes of both with the goal of creating KG Chatbots. 

\section{Statistics Analysis and Open Challenges}\label{sec:OpenChallenges}
\subsection{Statistics about used LLMs and KGs in Approaches}
Our analysis categorizes research papers based on the use of LLMs and KGs. Figure~\ref{fig:LLM-KG used} illustrates this, with the y-axis representing the frequency of appearances and the x-axis listing the specific KGs and LLMs. The data reveal that Freebase is the most commonly utilized KG in the reviewed literature. Regarding LLMs, Bert and GPT-3 emerge as the most frequently employed models. 

\subsection{Open Challenges}
We still need research to find techniques to incorporate knowledge reliably into the answers of LLMs and to go for smaller-sized LLMs without losing the capabilities of LLMs to reason on their input. One way seems to be to incorporate the knowledge from KGs reliably into the inference process of LLMs and to exclude the knowledge from the training data because these facts are then not needed anymore to be stored in the neural network of the LLMs \cite{DBLP:conf/nips/MengBAB22} increasing its number of parameters. Also, incorporating only code of a core fragment of a programming language or a query language like SPARQL \cite{Groppe2009CoreSPARQL} and XPath \cite{Groppe2006XpathSimplification,Groppe2006Rewriting} without redundant language constructs in the training data may reduce the number of needed parameters in LLMs. To reduce the number of needed parameters, future work may also consider generating code and queries in a limited set of programming and query languages and transforming the generated code to the requested programming and query languages. For example, automated transformations between XML and Semantic Web data format and query languages are well known \cite{Droop2007Translating,Droop2008XMLSWCloser,Droop2008Embedding,Groppe2008Embedding,Groppe2011Transforming}, and also reformulating queries based on a given transformation of the data~\cite{Groppe2006Reformulating}. Other techniques may consider only those queries in the training data, which can return a result and are hence satisfiable \cite{Groppe2006PrototypeSatisfiability,Groppe2008FilteringUnsatisfiableXPath,Groppe2009SWOBE,Groppe2006Rewriting}. Running LLMs with fewer parameters reduces the energy needed and, hence, the carbon footprint of running LLMs. We see a demand for a better separation of knowledge (to be managed in KGs in a condensed and reliable way) and natural language understanding (based on a well-chosen minimal training data set of high quality not addressing the knowledge of the given KG). How to retrieve this well-chosen minimal training data set is a non-trivial question and may open up a new research direction itself. With a clear separation of knowledge and natural language understanding, fine-tuning to specific domains may also become obsolete or at least can be reduced to a minimum effort. New applications of LLMs like Personal KG-enhanced LLMs, which can imitate the style of writing of each individual by fine-tuning from email and chat conversations and based on a Personal KG containing the (private) knowledge of the individual might become true.

Besides reducing the complexity of LLMs there might be another research direction on more complex architectures mimicking the working of the human brain even more towards the holy grail of artificial intelligence, the artificial general intelligence (AGI), to accomplish any intellectual task that human beings or animals can perform \cite{Shevlin19AGI}. In these architectures, LLMs or advanced variants might play only a minor role by being responsible for processing the input and/or verbalizing the conclusion of the overall process. KGs might still be a standard way for administrating knowledge in a reliable and space-efficient way in these architectures for AGI.

\section{Conclusion}\label{sec:Conclusions}
In conclusion, this comprehensive survey has significantly contributed to the understanding and advancement of the integration between LLMs and KGs. Through our meticulous exploration across various sections, we have addressed critical research questions, shedding light on how LLMs can enhance KGs in areas such as generating descriptive textual information, ontology generation, inconsistency detection, fact-checking, KG completion, and KG embedding. Our study also investigates how KGs can improve LLMs and explores the cooperation between LLMs and KGs. Specifically, it examines under-discussed areas in KG question answering, including query generation from text and multi-hop question generation and answering.

\begin{acks}
 The German Research Foundation funds this work under project number 490998901.
\end{acks}


\bibliographystyle{ACM-Reference-Format}
\bibliography{vldb}

\end{document}